\documentclass{article}

\PassOptionsToPackage{numbers}{natbib}

\usepackage[final]{neurips_2020}

\usepackage[utf8]{inputenc} 
\usepackage[T1]{fontenc}    
\usepackage{hyperref}       
\usepackage{amssymb}
\usepackage{url}            
\usepackage{booktabs}       
\usepackage{amsfonts}       
\usepackage{nicefrac}       
\usepackage{microtype}      
\usepackage{xcolor}         
\usepackage{times}
\usepackage{amsmath, amssymb}
\usepackage{mathtools}
\usepackage{latexsym}
\usepackage{todonotes}
\usepackage{comment}
\usepackage{tabto}
\usepackage{soul}
\usepackage{multicol,multirow}
\usepackage{graphicx}
\usepackage{verbatim}
\usepackage{authblk}

\def\BibTeX{{\rm B\kern-.05em{\sc i\kern-.025em b}\kern-.08em
    T\kern-.1667em\lower.7ex\hbox{E}\kern-.125emX}}
    
\usepackage{color, colortbl}    
\usepackage{amssymb}
\usepackage{algorithm}
\usepackage[noend]{algpseudocode}
\usepackage{xcolor}

\author[1]{\textbf{Sannara Ek}}
\author[1]{\textbf{Romain Rombourg}}
\author[1]{\textbf{François~Portet}}
\author[1]{\textbf{Philippe Lalanda}}

\affil[1]{Univ. Grenoble Alpes, CNRS, Inria, Grenoble INP, LIG, 38000 Grenoble, France  \qquad}

\title{Federated Self-Supervised Learning in Heterogeneous Settings: Limits of a Baseline Approach on HAR}

\begin{document}
\maketitle

\begin{abstract}
Federated Learning is a new machine learning paradigm dealing with distributed model learning on independent devices. One of the many advantages of federated learning is that training data stay on devices (such as smartphones), and only learned models are shared with a centralized server. In the case of supervised learning, labeling is entrusted to the clients. However,  acquiring such labels can be prohibitively expensive and error-prone for many tasks, such as human activity recognition. Hence, a wealth of data remains unlabelled and unexploited. Most existing federated learning approaches that focus mainly on supervised learning have mostly ignored this mass of unlabelled data.
Furthermore, it is unclear whether standard federated Learning approaches are suited to self-supervised learning. The few studies that have dealt with the problem have limited themselves to the favorable situation of homogeneous datasets. This work lays the groundwork for a reference evaluation of federated Learning with Semi-Supervised Learning in a realistic setting. We show that standard lightweight autoencoder and standard Federated Averaging fail to learn a robust representation for Human Activity Recognition with several realistic heterogeneous datasets. These findings advocate for a more intensive research effort in Federated Self Supervised Learning to exploit the mass of heterogeneous unlabelled data present on mobile devices. 
\end{abstract}

\section{Introduction}
Federated Learning (FL) was proposed by Google back in 2016 \cite{mcmahan2016communicationefficient} as a novel form of distributed meta-learning that allows specialized machine learning models to be trained directly on user devices without ever communicating user data. The baseline FL works as so: the server sends a model to multiple clients to individually train on their local data, then all trained personalized models are sent back to the server and aggregated to create a single generalized model that summarises the individual models acquired by every client to conclude a communication round. The process is then repeated until convergence. The method offers an efficient and flexible distributed learning machine learning paradigm that preserves user data privacy and seems well suited to pervasive computing applications.

\begin{figure}[!bt]
\centering
\includegraphics[scale=0.093]{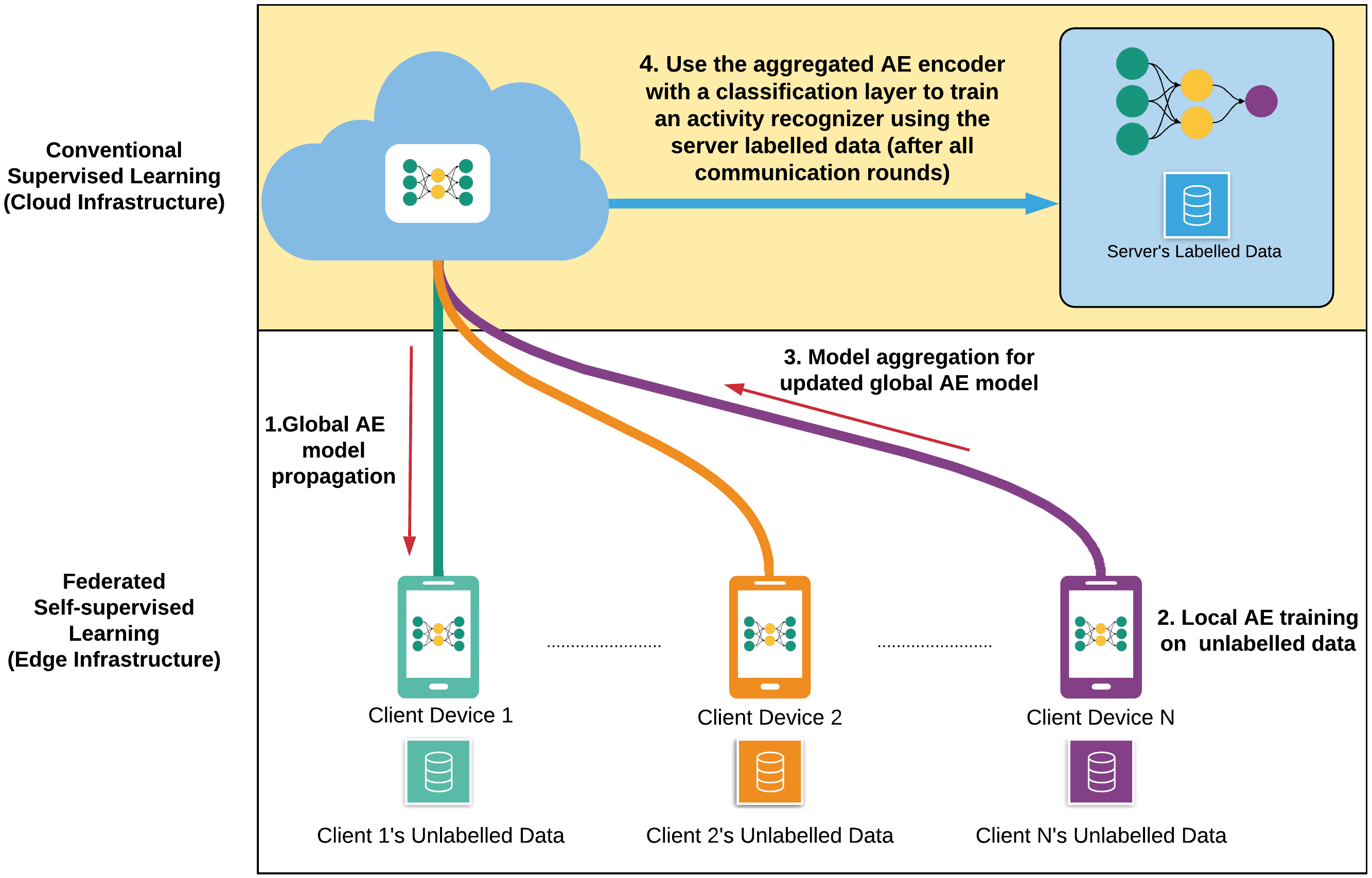}
\caption{Illustration of the federated learning process for the autoencoder. }
\label{fig:federatedLearning}
\end{figure}

This new paradigm has been adopted into some real-world applications, as exemplified by Google keyboard \cite{47586}. However, the applicability of FL with uncontrolled environments remains an open challenge \cite{kairouz2019advances}. Adopting FL into a pervasive computing domain, where connected smart devices are embedded into our environment, raises multiple challenges \cite{pervasiveTrend}. The most complex challenges are the scarcity of labeled data, the heterogeneity of users' data on mobile devices, and the low computation/memory constraints.
Most FL approaches have been studied with a supervised learning target. However, supervised learning relies on a large amount of accurately labeled data. Such labeling is a tedious process that is highly time-consuming for human experts. Furthermore, since data stays on users' devices in an FL setting, this would require the users to label their own data, which is not feasible most of the time. Hence, a large amount of data stays unlabelled on the users' devices. One way to leverage this wealth of unlabelled client data is to develop unsupervised learning methods that only require unlabelled raw data to learn useful representations \cite{zhang2020federated,selfsuper}. Another constraint when dealing with learning on devices is the often low computational resources available at the client level. This constraint implies that certain machine learning techniques or models that are too complex cannot be used in such FL scenarios, and lightweight learning techniques must be specifically designed. Finally, a major challenge throughout most environments of FL is data heterogeneity
\cite{li2020federated,wang2020federated}.
These constraints have been under-studied in the FL domain.  
Most recently proposed methods have either used a high resource-consuming architecture/schema, assumed the presence of large quantities of labeled data, or did not consider heterogeneity in their studies.

In this paper, we present the results of a first attempt to leverage the high amount of unlabelled user data for FL in a realistic heterogeneous setting. We use Human Activity Recognition on mobile devices as a representative pervasive computing application (low resource devices, heterogeneity, users daily self-labeling highly challenging). We implemented self-supervised feature learning using an AutoEncoder (AE) architecture \cite{pmlr-v27-baldi12a} for representation learning due to its low complexity. Such feature learning was performed through FL using the original  FedAvg algorithm \cite{mcmahan2016communicationefficient} which is also of low complexity. For the human activity classification task, 
we defined a realistic learning environment where clients have only unlabelled data while there is a small amount of non-private labeled data at the server level. 
The clients are created using multiple publicly available datasets that were collected from a range of different devices and distinct participants. Our study tries to answer the following research question: ``Can Autoencoders be used in a federated learning setting to learn efficient representations for human activity recognition from a heterogeneous set of unlabelled data?''

\section{Related Work}

There have been only a few studies on the notion of Self-supervised learning and federated learning. For instance, the study by \cite{zhang2020federated} has shown that a simple AE in federated learning can indeed provide performance improvements compared to a conventional learning approach. Additionally, the study has also shown that using more advanced feature-learning architecture allows even more profit in performance gains. While the study done in \cite{10.1145/3378679.3394530} that has as well investigated AEs applications onto a federated learning context shows that the AE does not supersede supervised learning.
Another study by \cite{selfsuper}, by designing a novel self-supervised contrasting learning network, has raised that self-supervised learning on a large scale can be well used for sensor modeling tasks. However, the method is based on a modified SimCLR for federated learning \cite{selfsuper}, which might be too complex for real mobile devices. 
Indeed, these study does not take into account the computing capabilities of the devices \cite{kairouz2019advances} since they propose big model architectures and complex configurations, e.g., dual-stream architecture, reliance on public datasets, and using large training batch-sizes. Furthermore, the evaluations were performed on isolated datasets that do not fully represent the challenges imposed by heterogeneity in the wild. To the best of our knowledge, learning with different datasets in FL has only been achieved by \cite{li2021survey}. However, this was solely for supervised learning. In our paper, we want to evaluate the capability of a lightweight auto-supervised model, namely Autoencoder, to be trained via federated learning on a large heterogeneous set of data. We could not find any work in the literature that has previously achieved this objective.

\section{Method}

\begin{figure}[!bt]
\centering
\includegraphics[scale=0.25]{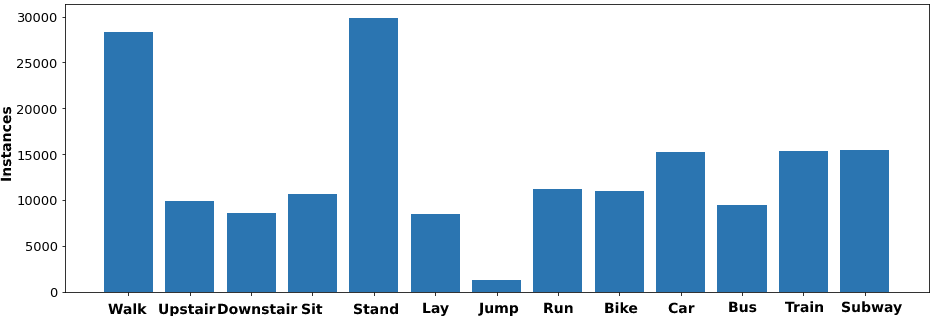}
\caption{Class distribution for the four datasets}
\label{fig:fedAvgAcc}
\end{figure}

 In order to best replicate a realistic non-iid environment, where system and statistical heterogeneity along with class/data imbalance plays a large role between clients, we combined four different publicly available datasets to have 80 diverse clients. We selected the following datasets: the {\bf UCI dataset} \cite{Anguita2013APD} -- the standard benchmark dataset in the HAR community -- which, in our experiments, represents five homogeneous clients, the \textbf{Heterogeneity Human Activity Recognition (HHAR)} dataset \cite{10.1145/2809695.2809718} which comprises 51 heterogeneous clients, the {\bf REALWORLD} dataset  \cite{realword} -- a large dataset with a high diversity of device positioning -- which adds 15 clients and finally the very large {\bf Sussex-Huawei Locomotion (SHL)} dataset \cite{8418369} that adds further nine clients. Each dataset brings sets of clients with unique properties to the combined dataset. More so, each of the datasets has its own set of activities, as shown in table \ref{fig:datasetProp} with only some overlapping. The combined dataset deals with a total of 13 unique activities: Walk (W), Upstairs (U), Downstairs (D), Sit (ST), Stand (SD), Lay (L), Jump (J), Run (R), Bike (BK), Car (C), Bus (BS), Train (T), Subway (SW). In terms of class distribution, shown in figure \ref{fig:fedAvgAcc}, ``Jump'' is the minority class while there is a large number of samples for the ``Stand'' and ``Walk'' classes.

From all the datasets, we used 3-axis accelerometer data with the 3-axis gyroscope data resampled at 50hz when necessary. We used a window-frame size of 128 (2.56s) with an overlap of 50\% (1.28s) for the 6 channels of each axis. No feature extraction was performed, and the data was preprocessed using channel-wise z-normalization.

\begin{table}[!htb]
\centering
\caption{Datasets properties}

\begin{tabular}{|l|c|c|c|}
\hline
Dataset  & Clients & Data Samples & Activities \\ 
\hline
UCI & 5 & 10,299  & 6 (ST,SD,W,U,D,L) \\ 
HHAR & 51& 85,567 & 6 (ST,SD,W,U,D,BK)\\
REALWORLD & 15 & 356,427 & 8 (ST,SD,W,U,D,J,L,R)  \\
SHL & 9 & 640,144 & 8 (ST,W,R,BK,C,BS,T,SW)  \\

\hline
\end{tabular}
\label{fig:datasetProp}
\end{table}

\begin{figure}[!bt]
\centering
\includegraphics[scale=0.23]{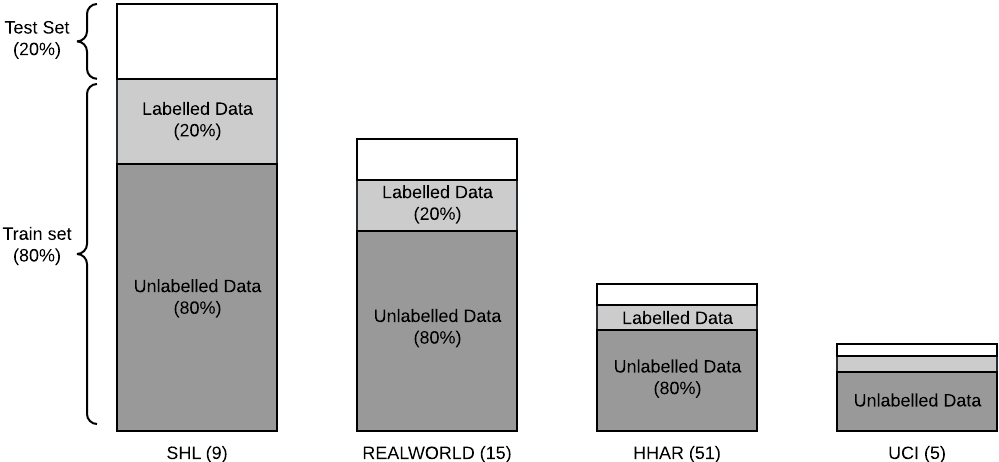}
\caption{Data Partitioning Methodology}
\label{fig:datapartition}
\end{figure}

The data partitioning was done as illustrated by Figure~\ref{fig:datapartition}. For each dataset, 20\% is left for testing the classifier at the server (Test-set), while the remaining 80\% is used for training (Train-set). To simulate an in-the-wild training environment, where clients' data are commonly unlabeled, we disregard 80\% of the labels of the train-set for unsupervised feature learning on client devices. On the other hand, the remaining 20\% train-set with labels are pooled together for supervised fine-tuning at the server. The final learning environment is then where we have 80 clients performing unsupervised feature learning in a federated manner and the server conducting supervised fine-tuning and testing.
We adapted the AE model architecture from \cite{varamin2018deep}. The chosen model comprises four 1d convolution layers for the encoder, each consisting of 32 filters with a kernel size of 5, leading to the latent space. The decoder uses the same settings but employs transposed convolution layer instead. Finally, 128 neurons were set as the size of the latent space. The AE model is trained in a federated learning way for 200 communication rounds with 5 local epoch for each client on a learning rate of 0.01 with a Stochastic Gradient Descent (SGD) optimizer. At the end of every local training, the clients evaluate the AE loss on their respective test-set. Additionally, after every server model aggregation, we evaluate the server model on the combined client test-set.

After the 200 communication rounds of training in a federated method, we take only the encoder of the AE and attach a dense layer of shape 32 that leads to the soft-max layer of shape 13 for classification. We used a learning rate of 0.00005 with an ADAM optimizer to train the model. The classifier is fine-tuned on the labeled data train-set for 200 epochs with class-weighted learning as a measure against the class imbalance. The classifier model evaluates on the combined test-set of all clients, where we then have the model evaluate separately on each set of the dataset's client and the combined test-set of all clients. The entire learning process is summarized in figure \ref{fig:federatedLearning}.

For comparative reasons in our study, we trained the model with two different settings using the same architecture and shape. First, in order to evaluate the benefits of the feature pre-training process of the AE, we train the AE entirely at the server level on an aggregated clients dataset (The combination of all clients unlabelled data). Secondly, intending to obtain quantitative analysis on the performance change between the FL way and the conventional way, we train a structurally identical model (four convolutional layers plus the dense and softmax layers) solely on the server labeled data with class-weighted learning.

The experiments were run on a Debian 4.19.132-1 version 10 having 256GB of RAM, 4x NVIDIA Quadro RTX 8000 48 GB GPU, and an Intel(R) Xeon(R) Silver 4210R CPU @ 2.40GHz for the CPU. For the federated learning experiment, 4 GPUs were used in parallel, while only 1 of the 4 was used for all the other cases.

\section{Results}

The overall results of the study are reported in Table~\ref{fig:resultbl2}. 
The AE learned using the FL method ({\bf FL+AE}) exhibits an F-score of 71.29\% on the combined test sets. On the classifier using {\bf conventional} learning -- which consisted in classical centralized supervised learning on the labeled data only. The conventional approach shows an F-score of 71.96\% on the combined test sets. It seems thus that either the FL learning or the AE are not able to benefit from the large set of unlabeled data. When the AE is trained in a centralized way ({\bf Conventional + AutoEncoder}), no improvement on the combined dataset is observed (69.04\%) while there is instead a slight decrease. It seems thus that the AE does not learn a representation helpful for the classification task.    

\begin{table}[!hb]
 \centering
  \caption{The performance of experiments on the combined datasets}

 \resizebox{\columnwidth}{!}{
 \begin{tabular}{|c||c||c|c|c|c|}
 \hline

 \multirow{2}{*}{Learning Method}  & \multicolumn{5}{c|}{ Datasets (Macro F-score) } \\ \cline{2-6}

  & Combined&UCI& HHAR& REALWORLD  & SHL \\ 
 \hline
 FL + Autoencoder & 71.29 & 73.51 & 82.84  & \textbf{77.98} & 68.81  \\
\hline
 Conventional & \textbf{71.96} &  \textbf{78.92} & \textbf{82.92}  & 76.84 & \textbf{70.44}  \\
 Conventional + Autoencoder & 69.04 & 73.06 & 78.58  & 75.48 & 66.52  \\

 \hline
 \end{tabular}
 }
 \label{fig:resultbl2}
 \end{table}

On the individual dataset, it is further apparent that \textbf{conventional} approach overall (with the exception of the REALWORLD dataset) has a slight performance edge overall, 73.51\%, 82.84\%, 77.98\% and 68.81\% respectively on the UCI, HHAR, REALWORLD, and SHL dataset. Herewith the \textbf{ FL + Autoencoder} approach, the scores are 73.51\% on UCI, 82.84\% on HHAR, 77.98\% on REALWORLD (The best amongst the three emulations), and 68.81\% on SHL dataset. Finally, on the \textbf{Conventional + Autoencoder} approach, we relatively see the lowest performance across all the datasets where we obtained 73.06\%, 78.58\%, 75.48\%, and 66.52\% respectively on the UCI, HHAR, REALWORLD, and SHL datasets. Overall, with marginal differences, no method seems superior to others.

\begin{figure*}[!bt]
\centering
\includegraphics[scale=0.37]{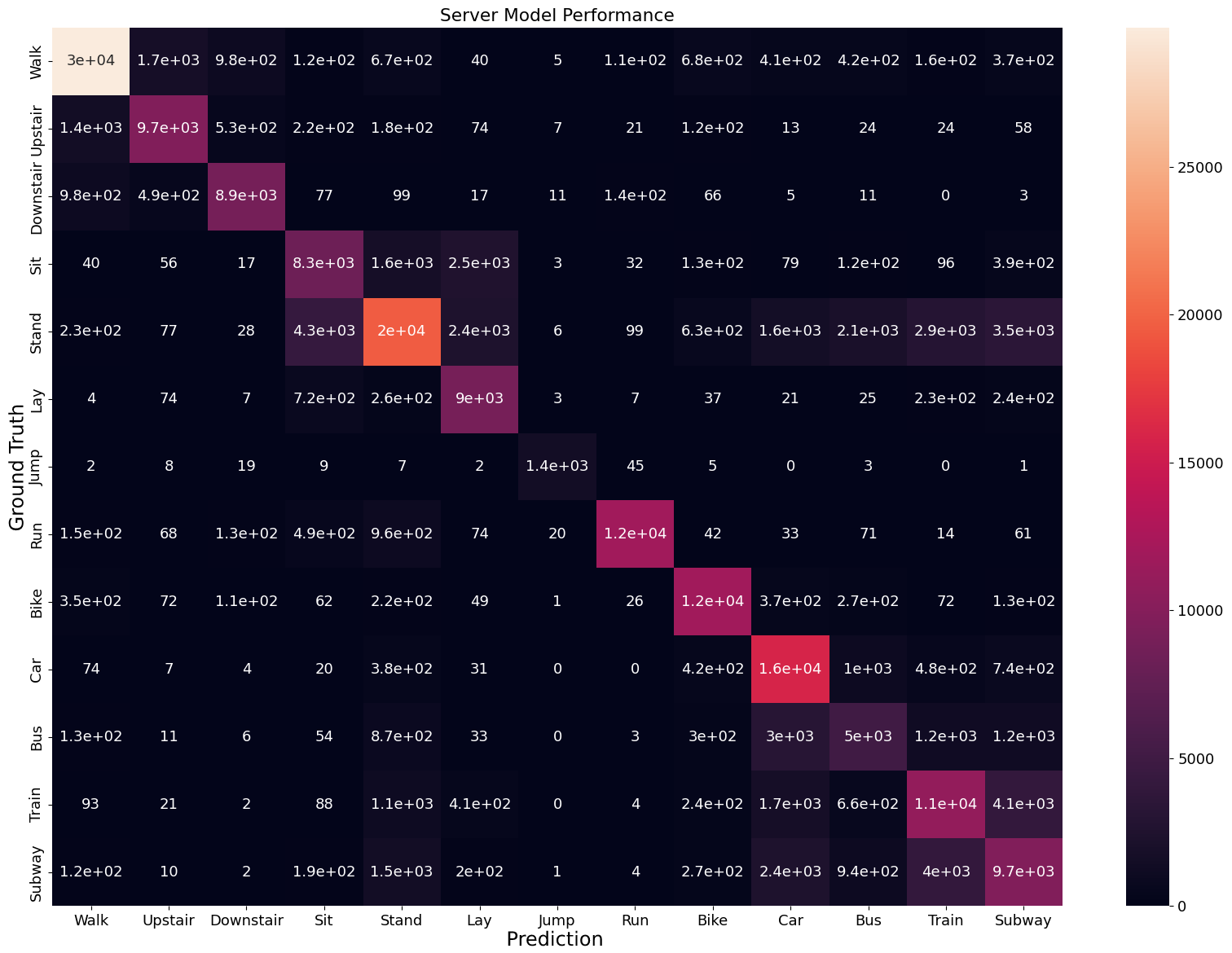}
\caption{Confusion Matrix of Classifier With Federated Learning Feature Learning}
\label{fig:confuMat}
\end{figure*}

\begin{table}[!htb]
\centering
\caption{Average storage footprint of local data and model}

\begin{tabular}{|l|c|}
\hline
Dataset/Model  & Size (MBs)   \\ 
\hline
UCI & ~3.87   \\ 
HHAR & ~3.15\\
REALWORLD & ~89.11   \\
SHL & ~266.73   \\
 \hline
Autoencoder &  ~0.38 \\

\hline
\end{tabular}
\label{fig:datassetSize}
\end{table}

\begin{figure}[!bt]
\centering
\includegraphics[scale=0.55]{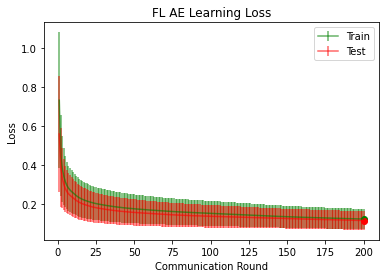}
\caption{Autoencoder Loss with FedAvg}
\label{fig:AEFL_LOSS}
\end{figure}

Regarding the FL learning, it can be seen from the learning curve presented by Figure~\ref{fig:AEFL_LOSS} that the training was regular. The averaged loss of the clients learning over 200 communication rounds shows that the learning is indeed progressing and able to converge. More so, a significant standard deviation can be observed on the client-side. This trait is likely due to the large number of dissimilar clients and the heterogeneity where the results show that such training with heterogeneous data is indeed challenging.  

Details relating to memory footprint, where the data are stored in an HDF5 format\cite{hickle}, is shown in table \ref{fig:datassetSize} where the average local data size of clients from the four different datasets vary significantly. Specifically, clients of UCI and HHAR contains around 3MB of data on average, while clients of REALWORLD have an average of 89.11 MBs, and SHL clients have an average size of 266.73 MBs. On the other hand, the communicated AE has a size of 0.38 MBs. Thus, we can establish that the communication cost for each client after 200 rounds of training in FL would be  76 MBs.

\begin{figure}[!bt]
\centering
\includegraphics[scale=0.23]{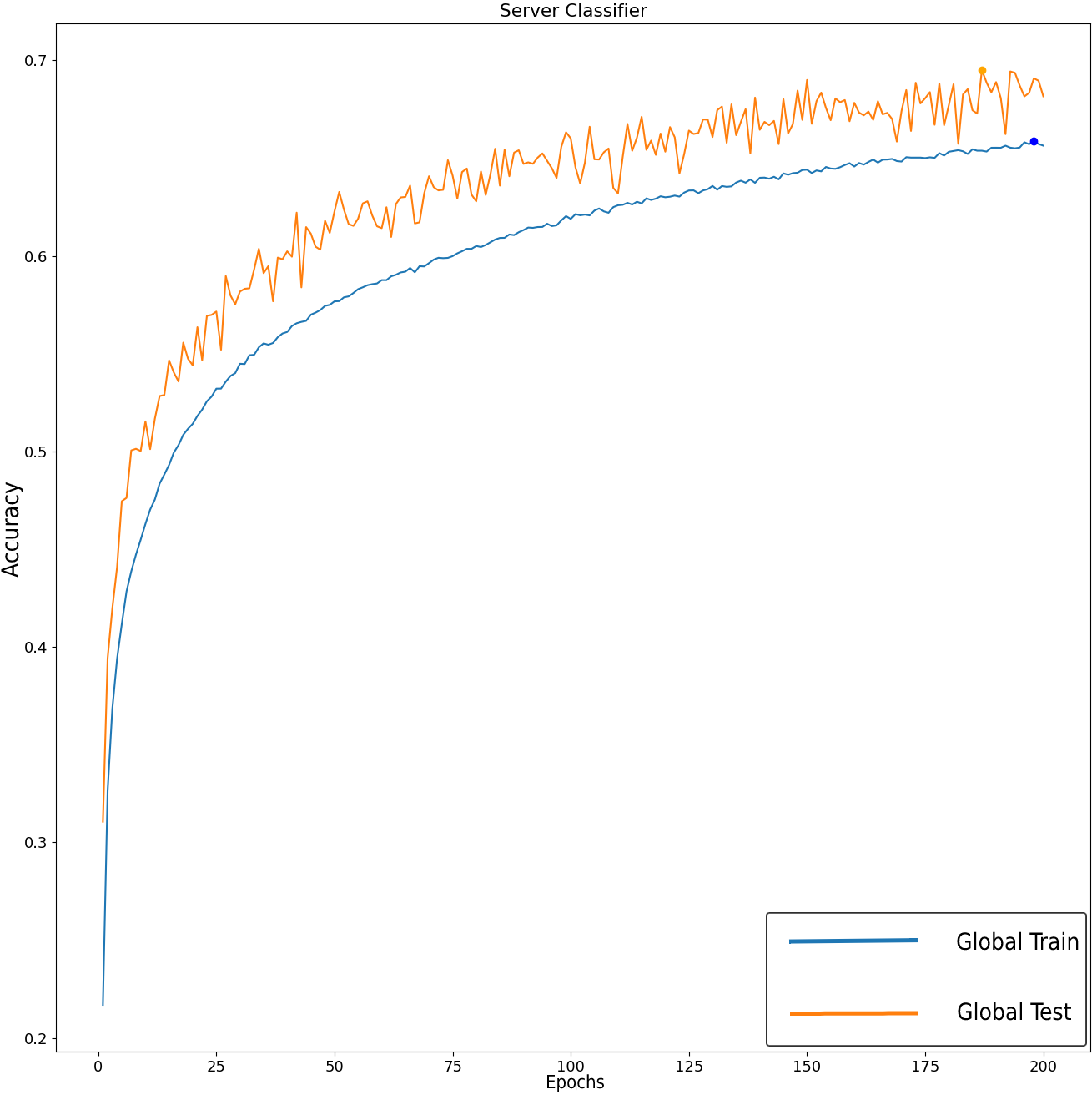}
\caption{Classifier Fine-Tuning on the Server}
\label{fig:FrozenLearn}
\end{figure}

\smallskip

\section{Discussion}
The presented combination of several distinctly dissimilar datasets has indeed presented a challenging learning problem. The results presented above have shown that even the model trained by the conventional/centralised learning method is unable to obtain good results compared to the studies that had trained individually and separately trained on only one dataset at a time \cite{IGNATOV2018915,ek:hal-03207411,yao2017deepsense,Sussex_challenge}.

We attribute the mitigated performance with the federated learning approach to several reasons. First, the nature of FedAvg's weighted averaging does not favor clients with little data. Clients with little data (from UCI and HHAR datasets) have a lower weight in the federated learning than clients with large data (REALWORLD and SHL datasets). This property explains why the performance of the UCI and HHAR datasets is much lower than the conventional approach. Hence, the global/server model is heavily driven mainly by the clients of the task-challenging SHL dataset (classification of transport mode) and partially by the heterogeneous REALWORLD dataset. To support this interpretation, a confusion matrix presented by Figure~\ref{fig:confuMat} reveals that the  classifier is able to correctly predict non-locomotion related activities e.g. ``Upstair'', ``Downstair'', ``Run'', ``Jump''. The model, however, was often confused in classifying between the different locomotion classes, namely, ``Bus'', ``Train'' and ``Subway''. This can be due to the signal z-normalization and the lack of context in the model representation.
Additionally, the mentioned locomotion classes are often confused with the ``Stand'' activity. This behavior of the model is quite understandable since, in public transport, a person is often standing while waiting at a stop. The classification task thus becomes reasonably tricky as the model must uniquely learn the subtle motion changes applied by the transportation vehicle to the ``stand'' activity.

\begin{figure}[!bt]
\centering
\includegraphics[scale=0.235]{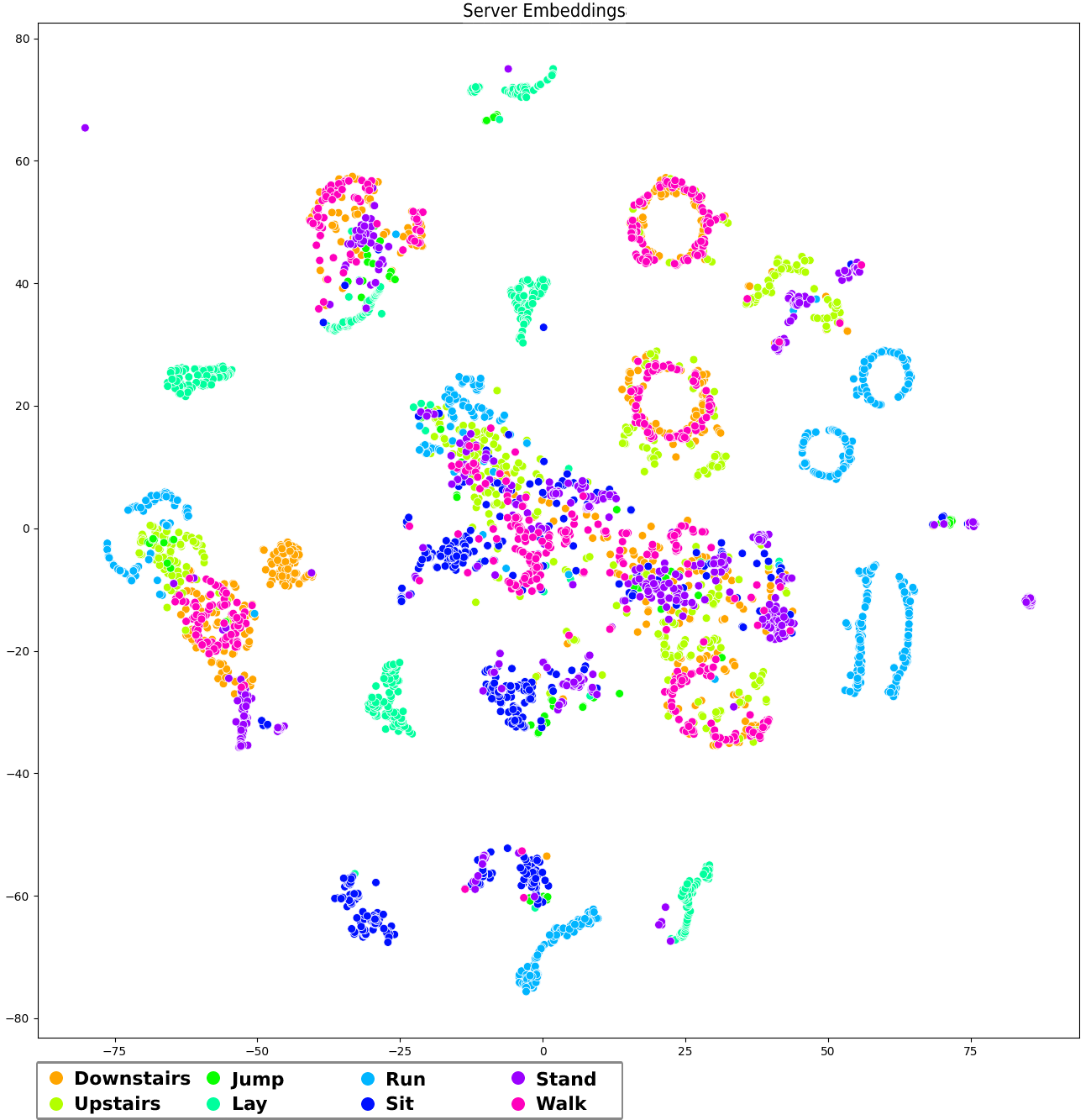}
\caption{Aggregated Autoencoder Embeddings on a REALWORLD client}
\label{fig:serverEmbed}
\end{figure}

Second, collaborative feature learning, as we have done by training the AE using FedAvg, presents an orientation and scaling issue when the server model is aggregated. Given the same data samples to multiple client models that have just finished local training to embed/project the data into their respective latent space, we would see that the projections across all the clients would be different. The problem has been well raised in \cite{zhang2020federated} and the study shows that a global set of rules of directions are needed in order to aggregate a suitable feature extractor that represents all client's data. The aggregation by FedAvg as done in our experiments, however, is a naive approach \cite{wang2020federated,ek:hal-03207411} and does not handle the mentioned problems. As shown in Figure~\ref{fig:serverEmbed} where the T-SNE dimensions reduced embeddings of the different classes are mostly intertwined with each other without distinct separations. The aggregation method would thus require more complex policies to properly merge the learned feature extractors of all the 80 clients in this study, favorably an aggregator that would preserve specialization and increase generalization.
On the other hand, the AE architecture is lightweight but very much decrepit. This claim can be further supported by the results seen in table \ref{fig:resultbl2}, where the conventional approach was able to outperform the conventional with AE approach. The feature-learning done with the AE is an expensive training process, in both the federated and conventional means, yet did not bring improvements. This lacking thus calls for newer feature learning architectures such as CPC \cite{oord2019representation}, SimCLR \cite{chen2020simple}, and Moco \cite{he2020momentum} which has exhibited improved feature extraction and may be the route for improved performance if hardware constraints are properly taken into account.

\section{Conclusion}
In this study, we leveraged the high amount of unlabeled data in mobile devices by studying how federated learning and lightweight self-supervised learning can be combined. We evaluated the approach on the Human Activity Recognition task using a realistic learning environment through the combination of several public datasets. For the models, we used an AE due to its low complexity and used FedAvg as the baseline server model aggregation technique. We showed that a simple auto-encoder is not an effective way to represent the input in neither a centralized nor a federated setting. However, federated learning can better compensate for this ineffective representation than centralized learning. This behavior shows that studying coupling FL and more advanced self-supervised learning method exhibits the potential to be very promising to handle unlabelled data, which ultimately allows more practical FL assimilation into pervasive applications.

\section{Acknowledgements}
This work has been partially funded by Naval Group, France and by MIAI@Grenoble Alpes (ANR-19-P3IA-0003) funded by the French program Investissement d’avenir.

\bibliographystyle{ieeetr} 
\bibliography{bibfile.bib}
\vspace{12pt}

\end{document}